\documentclass[runningheads]{llncs}
\usepackage{graphicx}
\usepackage[colorlinks=true,linkcolor=black,citecolor=blue,urlcolor=blue,]{hyperref}
\usepackage[doipre={doi:~}]{uri}

\usepackage{amssymb}
\usepackage{amsmath}
\usepackage{multirow}
\usepackage{subfigure}
\usepackage[table,xcdraw]{xcolor}
\usepackage{booktabs}
\usepackage[misc]{ifsym}
\begin{document}
%
\title{Iteratively Coupled Multiple Instance Learning from Instance to Bag Classifier for Whole Slide Image Classification}
\titlerunning{Iteratively Coupled Multiple Instance Learning}
%
\author{Hongyi Wang\inst{1} \and
Luyang Luo\inst{2} \and
Fang Wang\inst{3} \and
Ruofeng Tong\inst{1,4} \\
Yen-Wei Chen\inst{5,1} \and
Hongjie Hu\inst{3} \and
Lanfen Lin\inst{1}$^{(\scriptsize\textrm{\Letter})}$ \and
Hao Chen\inst{2,6}$^{(\scriptsize\textrm{\Letter})}$
}
\authorrunning{Wang et al.}
%
\institute{College of Computer Science and Technology, Zhejiang University, Hangzhou, China \and
Department of Computer Science and Engineering, The Hong Kong University of Science and Technology, Hong Kong, China \and
Department of Radiology, Sir Run Run Shaw Hospital, Hangzhou, China \and
Research Center for Healthcare Data Science, Zhejiang Lab, Hangzhou, China \and
College of Information Science and Engineering, Ritsumeikan University, Kusatsu, Japan \and
Department of Chemical and Biological Engineering, The Hong Kong University of Science and Technology, Hong Kong, China \\
\email{llf@zju.edu.cn}
\email{jhc@cse.ust.hk}
}

\maketitle              
\begin{abstract}
Whole Slide Image (WSI) classification remains a challenge due to their extremely high resolution and the absence of fine-grained labels. Presently, WSI classification is usually regarded as a Multiple Instance Learning (MIL) problem when only slide-level labels are available. MIL methods involve a patch embedding module and a bag-level classification module, but they are prohibitively expensive to be trained in an end-to-end manner. Therefore, existing methods usually train them separately, or directly skip the training of the embedder. Such schemes hinder the patch embedder's access to slide-level semantic labels, resulting in inconsistency within the entire MIL pipeline. To overcome this issue, we propose a novel framework called Iteratively Coupled MIL (ICMIL), which bridges the loss back-propagation process from the bag-level classifier to the patch embedder.
In ICMIL, we use category information in the bag-level classifier to guide the patch-level fine-tuning of the patch feature extractor. The refined embedder then generates better instance representations for achieving a more accurate bag-level classifier. By coupling the patch embedder and bag classifier at a low cost, our proposed framework enables information exchange between the two modules, benefiting the entire MIL classification model. We tested our framework on two datasets using three different backbones, and our experimental results demonstrate consistent performance improvements over state-of-the-art MIL methods. The code is available at: https://github.com/Dootmaan/ICMIL.

\end{abstract}
\section{Introduction}


Whole slide scanning is increasingly used in disease diagnosis and pathological research to visualize tissue samples. Compared to traditional microscope-based observation, whole slide scanning converts glass slides into gigapixel digital images that can be conveniently stored and analyzed. However, the high resolution of WSIs also makes their automated classification challenging \cite{lu2021ai}.
Patch-based classification is a common solution to this problem \cite{cvpr2016em,zhang2018whole,chen2022deep}. It predicts the slide-level label by first predicting the labels of small, tiled patches in a WSI. This approach allows for the direct application of existing image classification models, but requires additional patch-level labeling. Unfortunately, patch-level labeling by histopathology experts is expensive and time-consuming. Therefore, many weakly-supervised \cite{cvpr2016em,zhang2018whole} and semi-supervised \cite{cheng2022deep,chen2022deep} methods have been proposed to generate patch-level pseudo labels at a lower cost.
However, the lack of reliable supervision directly hinders the performance of these methods, and serious class-imbalance problems could arise, as tumor patches may only account for a small portion of the entire WSI \cite{dsmil}.

\begin{figure}[t]
\includegraphics[width=\textwidth]{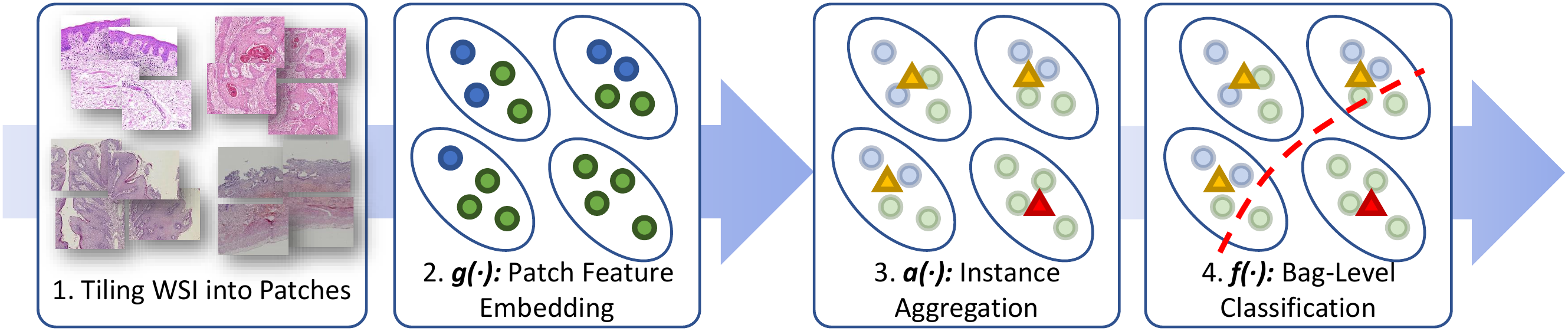}
\caption{The typical pipeline of traditional MIL methods on WSIs. } 
\label{MIL}
\end{figure}

\begin{figure}[t]
\center
\includegraphics[width=0.97\textwidth]{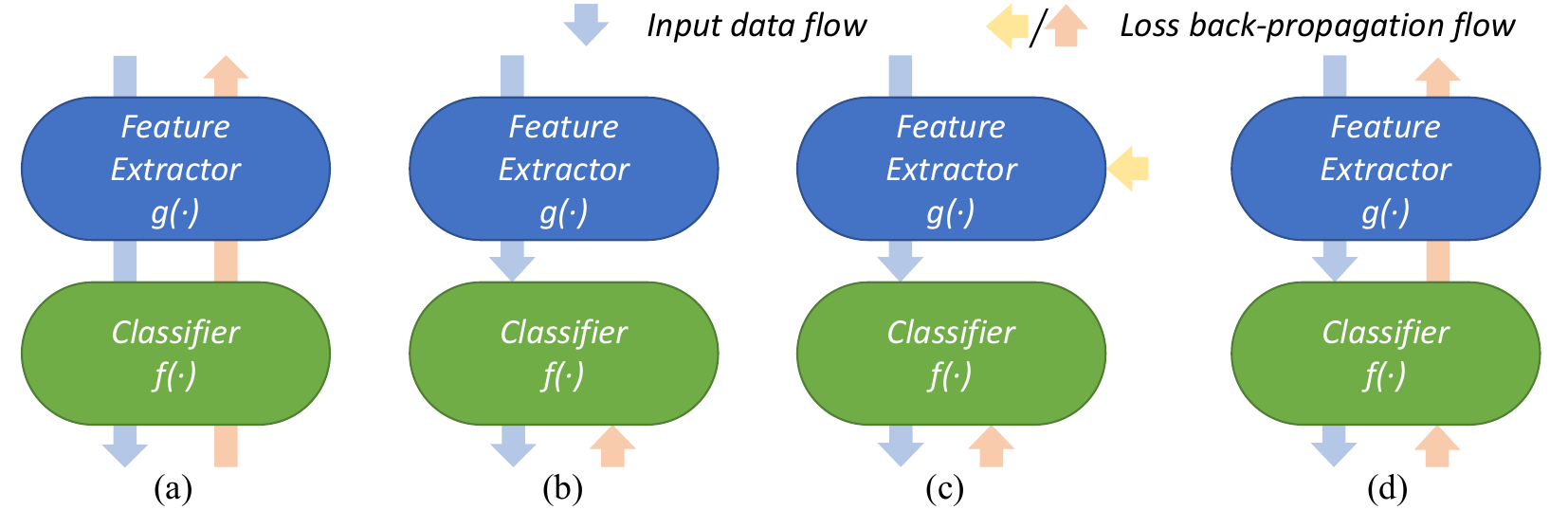}
\caption{Comparison between ICMIL and existing methods. (a) Ordinary end-to-end classification pipeline. (b) MIL methods that use fixed pre-trained ResNet50 as $g(\cdot)$. (c) MIL methods that introduce extra self-supervised fine-tuning of $g(\cdot)$. (d) Our proposed ICMIL which can bridge the loss back-propagation process from $f(\cdot)$ to $g(\cdot)$ by iteratively coupling them during training. }
\label{Difference}
\end{figure}

In contrast, MIL-based methods have become increasingly preferred due to their only demand for slide-level labels \cite{maron1997framework}. The typical pipeline of MIL methods is shown in Fig.~\ref{MIL}, where WSIs are treated as bags, and tiled patches are considered as instances. The aim is to predict whether there are positive instances, such as tumor patches, in a bag, and if so, the bag is considered positive as well. In practice, a fixed ImageNet pre-trained feature extractor $g(\cdot)$ is usually used to convert the tiled patches in a WSI into feature maps due to limited GPU memory. These instance features are then aggregated by $a(\cdot)$ into a slide-level feature vector to be sent to the bag-level classifier $f(\cdot)$ for MIL training.
Due to the high computational cost, end-to-end training of the feature extractor and bag classifier is prohibitive, especially for high-resolution WSIs. As a result, many methods focus solely on improving $a(\cdot)$ or $f(\cdot)$, leaving $g(\cdot)$ untrained on the WSI dataset (as shown in Fig.\ref{Difference}(b)). However, the domain shift between WSI and natural images may lead to sub-optimal representations, so recently there have been methods proposed to fine-tune $g(\cdot)$ using self-supervised techniques \cite{srinidhi2022self,dsmil,hipt} or weakly-supervised techniques \cite{liu2022multiple,wang2020ud,jin2023label} (as shown in Fig.\ref{Difference}(c)). Nevertheless, since these two processes are still trained separately with different supervision signals, they lack joint optimization and may still leads to inconsistency within the entire MIL pipeline.




To address the challenges mentioned above, we propose a novel MIL framework called ICMIL, which can iteratively couple the patch feature embedding process with the bag-level classification process to enhance the effectiveness of MIL training (as illustrated in Fig.~\ref{Difference}(d)). Unlike previous works that mainly focused on designing sophisticated instance aggregators $a(\cdot)$ \cite{dsmil,shao2021transmil,lu2021smile} or bag classifiers $f(\cdot)$ \cite{abmil,clam,dtfdmil}, we aim to bridge the loss back-propagation process from $f(\cdot)$ to $g(\cdot)$ to improve $g(\cdot)$'s ability to perceive slide-level labels. Specifically, we propose to use the bag-level classifier $f(\cdot)$ to initialize an instance-level classifier $f'(\cdot)$, enabling $f(\cdot)$ to use the category knowledge learned from bag-level features to determine each instance's category.
In this regard, we further propose a teacher-student \cite{hinton2015distilling} approach to effectively generate pseudo labels and simultaneously fine-tune $g(\cdot)$. After fine-tuning, the domain shift problem is alleviated in $g(\cdot)$, leading to better patch representations. The new representations can be used to train a better bag-level classifier in return for the next round of iteration.


In summary, our contributions are: (1) We propose ICMIL which bridges the loss propagation from the bag classifier to the patch embedder by iteratively coupling them during training. This framework fine-tunes the patch embedder based on the bag-level classifier, and the refined embeddings, in turn, help train a more accurate bag-level classifier. (2) We propose a teacher-student approach to achieve effective and robust knowledge transfer from the bag-level classifier $f(\cdot)$ to the instance-level representation embedder $g(\cdot)$. (3) We conduct extensive experiments on two datasets using three different backbones and demonstrate the effectiveness of our proposed framework.

\section{Methodology}
\subsection{Iterative Coupling of Embedder and Bag Classifier in ICMIL}
\begin{figure}[t]
\center
\includegraphics[width=0.83\textwidth]{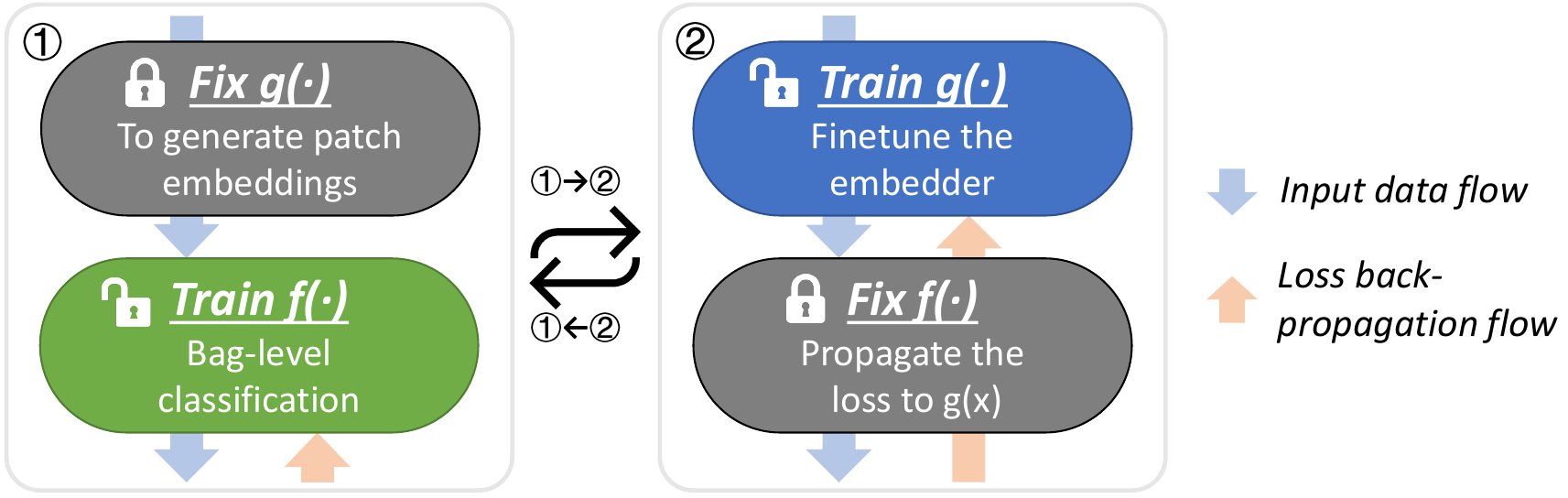}
\caption{The core idea of ICMIL: iteratively, \textcircled{1} fix the embedder $g(\cdot)$ and train the bag classifier $f(\cdot)$, \textcircled{2} fix the classifier $f(\cdot)$ and fine-tune the instance embedder $g(\cdot)$.}
\label{general_idea}
\end{figure}

The general idea of ICMIL is shown in Fig.~\ref{general_idea}, which is inspired by the Expectation-Maximization (EM) algorithm. EM has been used with MIL in some previous works \cite{luo2020weakly,liu2022multiple,wang2017instance}, but it was only treated as an assisting tool for aiding the training of either $g(\cdot)$ or $f(\cdot)$ in the traditional MIL pipelines. In contrast, we are the first to consider the optimization of the entire MIL pipeline as an EM alike problem, utilizing EM for coupling $g(\cdot)$ and $f(\cdot)$ together iteratively.
To begin with, we first employ a traditional approach to train a bag-level classifier $f(\cdot)$ on a given dataset, with patch embeddings generated by a fixed ResNet50 \cite{resnet} pre-trained on ImageNet \cite{imagenet} (step \textcircled{1} in Fig.\ref{general_idea}). Subsequently, this $f(\cdot)$ is considered as the initialization of a hidden instance classifer $f'(\cdot)$, generating pseudo-labels for each instance-level representation. This operation is feasible when the bag-level representations aggregated by $a(\cdot)$ are in the same hidden space as the instance representations, and most aggregation methods (e.g., max pooling, attention-based) satisfy this condition since they essentially make linear combinations of instance-level representations. 

Next, we freeze the weights of $f(\cdot)$ and fine-tune $g(\cdot)$ with the generated pseudo-labels (step \textcircled{2} in Fig.~\ref{general_idea}), of which the detailed implementation is presented in Section~\ref{tsfinetuning}.
After this, $g(\cdot)$ is fine-tuned for the specific WSI dataset, which allows it to generate improved representations for each instance, thereby enhancing the performance of $f(\cdot)$. Moreover, with a better $f(\cdot)$, we can use the iterative coupling technique again, resulting in further performance gains and mitigation to the distribution inconsistencies between instance- and bag-level embeddings.

\subsection{Instance Aggregation Method in ICMIL}
Although most instance aggregators are compatible with ICMIL, they still have an impact on the efficiency and effectiveness of ICMIL. In addition to that $a(\cdot)$ has to project the bag representations to the same hidden space as the instance representations, it also should avoid being over-complicated. Otherwise, $a(\cdot)$ may lead to larger difference between the decision boundaries of bag-level classifer $f(\cdot)$ and instance-level classifier $f'(\cdot)$, which may cause ICMIL taking more time to converge.

Therefore, in our experiments, we choose to use the attention-based instance aggregation method \cite{abmil} which has been widely used in many of the existing MIL frameworks \cite{abmil,clam,dtfdmil}. For a bag that contains $K$ instances, attention-based aggregation method firstly learns an attention score for each instance. Then, the aggregated bag-level representation $H$ is defined as:

\begin{equation}
    H=\sum_{k=1}^Ka_kh_k,
\end{equation}
\begin{equation}
    a_k=\frac{{\rm exp}\{\omega^T(tanh(V_1h_k)\odot sigm(V_2h_k))\}}{\sum_{j=1}^K{\rm exp}\{ \omega^T(tanh(V_1h_j)\odot sigm(V_2h_j))\}},
\end{equation}

\noindent where $a_k$ is the attention score for the $k$-th instance $h_k$ in the bag. Obviously, $H$ and $h_k$ remains in the same hidden space, satisfying the prerequisite of ICMIL. 





\begin{figure}[t]
\center
\includegraphics[width=\textwidth]{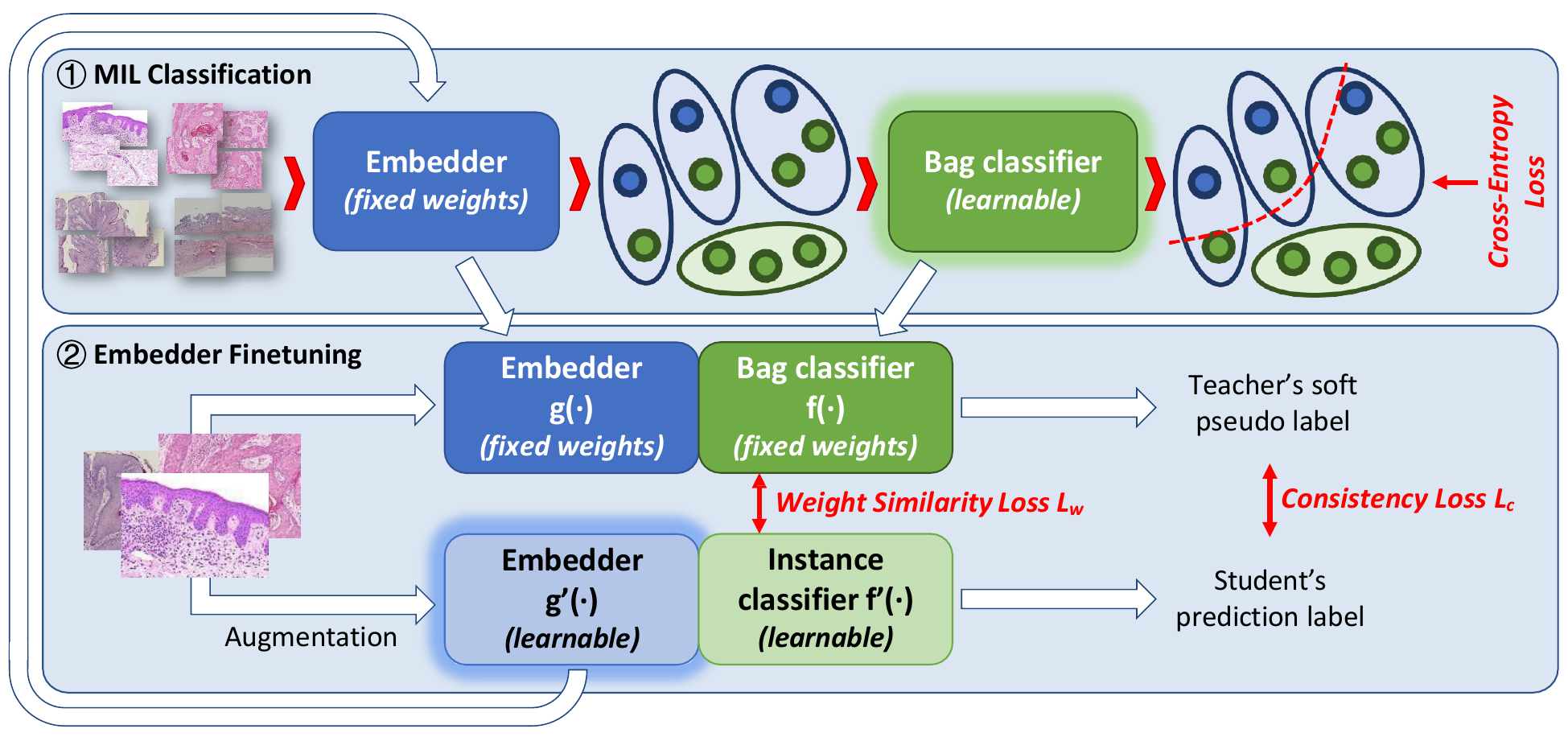}
\caption{A schematic view of the proposed teacher-student alike model for label propagation from $f(\cdot)$ to $g(\cdot)$ (mainly in step \textcircled{2}), and its position in ICMIL pipeline.} 
\label{knowledge_distillation}
\end{figure}

\subsection{Label Propagation from Bag Classifier to Embedder}
\label{tsfinetuning}
We propose a novel teacher-student model for accurate and robust label propagation from $f(\cdot)$ to $g(\cdot)$. The model's architecture is depicted in Fig.~\ref{knowledge_distillation}. In contrast to the conventional approach of generating all pseudo labels and retraining $g(\cdot)$ from scratch, our proposed method can simultaneously process the pseudo label generation and $g(\cdot)$ fine-tuning tasks, making it more flexible. Moreover, incorporating augmented inputs in the training process allows for the better utilization of supervision signals, resulting in a more robust $g(\cdot)$. We also introduce a learnable $f'(\cdot)$ to self-adaptively modifying the instance-level decision boundary for more effective fine-tuning of the embedder. 

Specifically, we freeze the weights of $g(\cdot)$ and $f(\cdot)$ and set them as the teacher. We then train a student patch embedding network, $g'(\cdot)$, to learn category knowledge from the teacher. For a given patch input $x$, the teacher generates the corresponding pseudo label, while the student receives an augmented image $x'$ and attempts to generate a similar prediction to that of the teacher through a consistency loss $L_c$. This loss function is defined as:

\begin{equation}
    L_c=\sum_{c=1}^{C} \left[f(x)_c log\left( \frac{f(x)_c}{f'(x')_c}\right)\right],
\end{equation}

\noindent where $f(\cdot)$ and $f'(\cdot)$ are teacher classifer and student classifier respectively, $f(\cdot)_c$ indicates the c-th channel of $f(\cdot)$, and $C$ is the total number of channels. 

Additionally, during training, a learnable instance-level classifier is used on the student to back-propagate the gradients to $g'(\cdot)$. The initial weights of $f'(\cdot)$ are the same as those of $f(\cdot)$, as the differences in the instance- and bag-level classification boundaries is expected to be minor. To make $f'(\cdot)$ not so different from $f(\cdot)$ during training, a weight similarity loss, $L_w$, is further imposed to constrain it by drawing closer their each layer's outputs under the same input. By applying $L_w$, the patch embeddings from $g'(\cdot)$ can still suit the bag-level classification task well, rather than being tailored solely for the instance-level classifier $f'(\cdot)$. $L_w$ is defined as:

\begin{equation}
    L_w=\sum_{l=1}^{L} \sum_{c=1}^{C} \left[f(x)_c^l log\left( \frac{f(x)_c^l}{f'(x)_c^l}\right)\right],
\end{equation}

\noindent where $f(\cdot)_c^l$ indicates the c-th channel of l-th layer's output in $f(\cdot)$. The overall loss function for this step is $L_c+\alpha L_w$, with $\alpha$ set to 0.5 in our experiments.

\section{Experiments}
\subsection{Datasets}
Our experiments utilized two datasets, with the first being the publicly available breast cancer dataset, Camelyon16 \cite{camelyon16}. This dataset consists of a total of 399 WSIs, with 159 normal and 111 metastasis WSIs for the training set, and the remaining 129 for test. Although patch-level labels are officially provided in Camelyon16, they were not used in our experiments. 

The second dataset is a private hepatocellular carcinoma (HCC) dataset collected from Sir Run Run Shaw Hospital, Hangzhou, China. This dataset comprises a total of 1140 valid tumor WSIs scanned at 40$\times$ magnification, and the objective is to identify the severity of each case based on the Edmondson-Steiner (ES) grading. The ground truth labels are binary classes of low risk and high risk, which were provided by experienced pathologists. 
\subsection{Implementation Details}
For Camelyon16, we tiled the WSIs into 256$\times$256 patches on 20$\times$ magnification using the official code of \cite{dtfdmil}, while for the HCC dataset the patches are 384$\times$384 on 40$\times$ magnification following the pathologists' advice. For both datasets, we used an ImageNet pre-trained ResNet50 to initialize $g(\cdot)$. The instance embedding process was the same of \cite{clam}, which means for each patch, it would be firstly embedded into a 1024-dimension vector, and then be projected to a 512-dimension hidden space for further bag-level training. For the training of bag classifier $f(\cdot)$, we used an initial learning rate of 2e-4 with Adam \cite{adam} optimizer for 200 epochs with batch size being 1. Camelyon16 results are reported on the official test split, while the HCC dataset used a 7:1:2 split for training, validation and test. For the training of patch embedder $g(\cdot)$, we used an initial learning rate of 1e-5 with Adam \cite{adam} optimizer with the batch size being 100. Three metrics were used for evaluation. Namely, area under curve (AUC), F1 score, and slide-level accuracy (Acc). Experiments were all conducted on a Nvidia Tesla M40 (12GB).

\begin{table}[t]
\renewcommand\tabcolsep{3pt}
\renewcommand\arraystretch{0.93}
\caption{Results of ablation studies on Camelyon16 with AB-MIL.}
\center
\label{ablation_study}
\subtable[Ablation study on the ICMIL iteration times]{
\begin{tabular}{l
>{\columncolor[HTML]{F4F9FF}}c 
>{\columncolor[HTML]{ECF4FF}}c 
>{\columncolor[HTML]{ECF4FF}}c 
>{\columncolor[HTML]{DAE8FC}}c 
>{\columncolor[HTML]{DAE8FC}}c 
>{\columncolor[HTML]{CDE1FF}}c 
>{\columncolor[HTML]{CDE1FF}}c }
\hline
ICMIL Iterations & 0     & 0.5 & 1     & 1.5  & 2     & 2.5  & 3    \\ \hline
AUC              & 85.4 & 88.8    & 90.0 & 89.7 & 90.5 & 90.4 & 90.0 \\
F1               & 78.0 & 79.4    & 80.5 & 80.1 & 82.0 & 80.7 & 81.7 \\
Acc              & 84.5 & 85.0    & 86.6 & 86.0 & 85.8 & 86.9 & 86.6    \\ \hline
\end{tabular}
}
\subtable[Loss Propagation]{
\begin{tabular}{ccc}
\hline
Method & Naïve & Ours \\ \hline
AUC    & 88.5     & 90.0      \\
F1     & 78.8     & 80.5      \\
Acc    & 83.9     & 86.6      \\ \hline
\end{tabular}
}
\end{table}



\begin{table}[t]
\renewcommand\tabcolsep{1.8pt}
\renewcommand\arraystretch{0.94}
\caption{Comparison with other methods on Camelyon16 and HCC datasets, where $^{\dagger}$ indicates the corresponding Camelyon16 results are cited from \cite{dtfdmil}. Best results are in bold, while the second best ones are underlined.}
\center
\label{experimental_results}
\begin{tabular}{cccccccccc}
\hline
\multirow{2}{*}{Method} & \multicolumn{3}{c}{Loss Propagation}                         & \multicolumn{3}{c}{Camelyon16} & \multicolumn{3}{c}{HCC} \\ \cmidrule(l){2-4} \cmidrule(l){5-7}  \cmidrule(l){8-10} 
                        & $g(\cdot)$                      & $f(\cdot)$ & $f$$\rightarrow$$g$                 & AUC(\%)      & F1(\%)       & Acc(\%)      & AUC(\%)    & F1(\%)     & Acc(\%)   \\ \hline
Mean Pooling            &                           & \checkmark     &  & 60.3    & 44.1    & 70.1    & 76.4  & 83.1  & 73.7 \\
Max Pooling             &                           & \checkmark     &  & 79.5    & 70.6    & 80.3    &  80.1      &  84.3      & 76.8      \\
RNN-MIL$^{\dagger}$ \cite{rnnmil}                 &                           &   \checkmark   &  & 87.5    & 79.8    & 84.4    & 79.4       &  84.1      & 75.5      \\
AB-MIL$^{\dagger}$ \cite{abmil}                 &                           &  \checkmark    &  & 85.4    & 78.0    & 84.5    & 81.2  & 86.0  & 78.1 \\
DS-MIL$^{\dagger}$ \cite{dsmil}                 &   \checkmark    &  \checkmark    &  & 89.9    & 81.5    & 85.6    & 86.1       & 86.6       & 81.4      \\
CLAM-SB$^{\dagger}$  \cite{clam}                &                           & \checkmark     &  & 87.1    & 77.5    & 83.7    &  82.1      & 84.3       & 77.1      \\
CLAM-MB$^{\dagger}$ \cite{clam}                &                           & \checkmark      & & 87.8    & 77.4    & 82.3    &  81.7      & 83.7       & 76.3      \\
TransMIL$^{\dagger}$ \cite{shao2021transmil}               &                           & \checkmark     &  & 90.6    & 79.7    & 85.8    & 81.2       & 84.4       & 76.7      \\
DTFD-MIL \cite{dtfdmil}               &                           &   \checkmark   &  & \underline{93.2}    & \underline{84.9}    & \underline{89.0}    & 83.0  & 85.5  & 78.1 \\ \hline
\vspace{-0.5mm}$\rm \mathop{Ours}\limits_{(w/\ Max\ Pooling)}$ & \checkmark &  \checkmark    & \checkmark &   $\mathop{85.2}\limits_{(+5.7)}$      & $\mathop{74.7}\limits_{(+4.1)}$         & $\mathop{81.9}\limits_{(+1.6)}$    & $\mathop{86.6}\limits_{(+6.5)}$  & $\mathop{87.3}\limits_{(+3.0)}$     & $\mathop{82.0}\limits_{(+5.2)}$  \\
\vspace{-0.5mm}$\rm \mathop{Ours}\limits_{(w/\ AB-MIL)}$    & \checkmark & \checkmark     & \checkmark & $\mathop{90.0}\limits_{(+4.6)}$   & $\mathop{80.5}\limits_{(+2.5)}$    & $\mathop{86.6}\limits_{(+2.1)}$  & $\mathop{87.1}\limits_{\underline{(+5.9)}}$        &  $\mathop{88.3}\limits_{\underline{(+2.3)}}$       & $\mathop{83.3}\limits_{\underline{(+5.2)}}$       \\
\vspace{-0.5mm}$\rm \mathop{Ours}\limits_{(w/\ DTFD-MIL)}$   & \checkmark &  \checkmark    & \checkmark 
&  $\boldsymbol{\mathop{93.7}\limits_{(+0.5)}}$      & $\boldsymbol{\mathop{87.0}\limits_{(+2.1)}}$   & $\boldsymbol{\mathop{90.6}\limits_{(+1.6)}}$  & $\boldsymbol{\mathop{87.7}\limits_{(+4.7)}}$     &  $\boldsymbol{\mathop{89.1}\limits_{(+3.6)}}$      &  $\boldsymbol{\mathop{83.5}\limits_{(+5.4)}}$     \\ \hline
\end{tabular}
\end{table}

\begin{figure}[t]
\includegraphics[width=\textwidth]{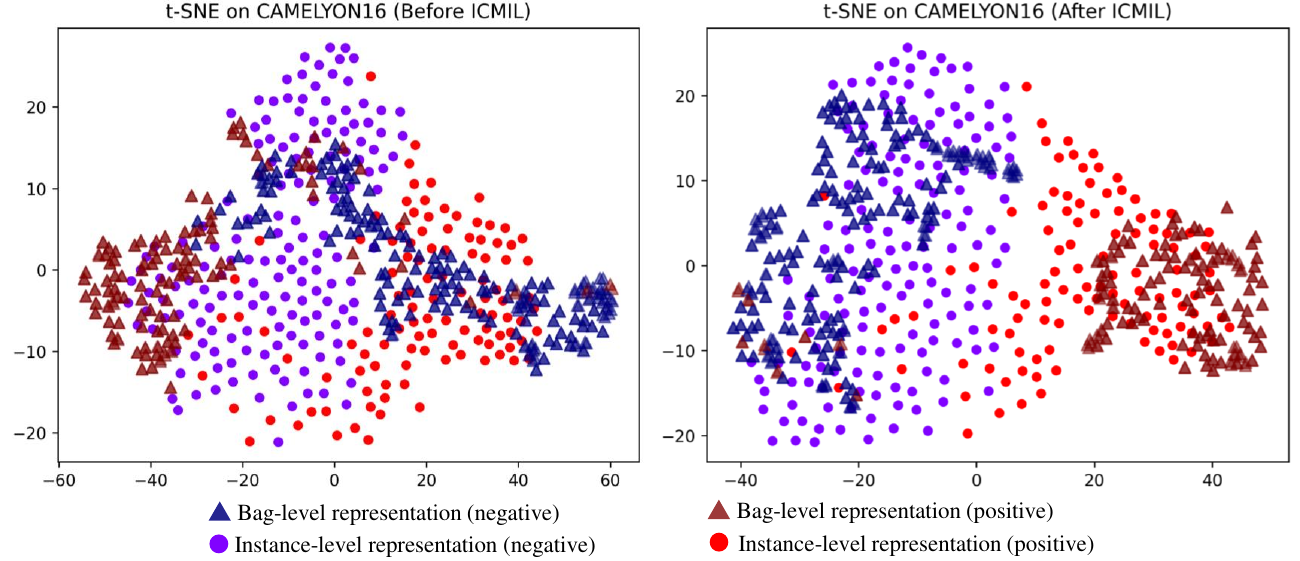}
\caption{Visualization of the instance- and bag-level representations before and after ICMIL training. We sample one instance from one bag w/ Max Pooling. Only one iteration of ICMIL is used to achieve the right figure.} 
\label{visualization}
\end{figure}

\subsection{Experimental Results}
\subsubsection{Ablation Study.}
The results of ablation studies are presented in Table~\ref{ablation_study}. From Table~\ref{ablation_study}(a), we can learn that as the number of ICMIL iteration increases, the performance will also go up until reaching a stable point. Since the number of instances is very large in WSI datasets, we empirically recommend to choose to run ICMIL one iteration for fine-tuning $g(\cdot)$ to achieve the balance between performance gain and time consumption. From Table~\ref{ablation_study}(b), it is shown that our teacher-student-based method outperforms the naïve ``pseudo label generation" method for fine-tuning $g(\cdot)$, which demontrates the effectiveness of introducing the learnable instance-level classifier $f'(\cdot)$.


\subsubsection{Comparison with Other Methods.}
Experimental results are presented in Table~\ref{experimental_results}. As shown, our ICMIL framework consistently improves the performance of three different MIL baselines (i.e., Max Pool, AB-MIL, and DTFD-MIL), demonstrating the effectiveness of bridging the loss back-propagation from bag calssifier to embedder. It proves that a more suitable patch embedding can greatly enhance the overall MIL classification framework. When used with the state-of-the-art MIL method DTFD-MIL, ICMIL further increases its performance on Camelyon16 by 0.5\% AUC, 2.1\% F1, and 1.6\% Acc.
Results on the HCC dataset also proves the effectiveness of ICMIL, despite the minor difference on the relative performance of baseline methods. Mean Pooling performs better on this dataset due to the large area of tumor in the WSIs (about 60\% patches are tumor patches), which mitigates the impact of average pooling on instances. Also, the performance differences among different vanilla MIL methods tends to be smaller on this dataset since risk grading is a harder task than Camelyon16. In this situation, the quality of instance representations plays a crucial role in generating more separable bag-level representations. 
As a result, after applying ICMIL on the MIL baselines, these methods all gain great performance boost on the HCC dataset.


Furthermore, Fig.~\ref{visualization} displays the instance-level and bag-level representations of Camelyon16 dataset before and after applying ICMIL on AB-MIL backbone. The results indicate that one iteration of $g(\cdot)$ fine-tuning in ICMIL significantly improves the instance-level representations, leading to a better aggregated bag-level representation naturally. Besides, the bag-level representations are also more closely aligned with the instance representations, proving that ICMIL can reduce the inconsistencies between $g(\cdot)$ and $f(\cdot)$ by coupling them together for training, resulting in a better separability.

\section{Conclusion}
In this work, we propose ICMIL, a novel framework that iteratively couples the feature extraction and bag classification stages to improve the accuracy of MIL models. ICMIL leverages the category knowledge in the bag classifier as pseudo supervision for embedder fine-tuning, bridging the loss propagation from classifier to embedder. We also design a two-stream model to efficiently facilitate such knowledge transfer in ICMIL. The fine-tuned patch embedder can provide more accurate instance embeddings, in return benefiting the bag classifier. The experimental results show that our method brings consistent improvement to existing MIL backbones.

\subsubsection{Acknowledgements} This work was supported by the National Key Research and Development Project (No. 2022YFC2504605), National Natural Science Foundation of China (No. 62202403) and Hong Kong Innovation and Technology Fund (No. PRP/034/22FX). It was also supported in part by the Grant in Aid for Scientific Research from the Japanese Ministry for Education, Science, Culture and Sports (MEXT) under the Grant No. 20KK0234, 21H03470.
%
%
%
\bibliographystyle{splncs04}
\bibliography{paper322}
%




\end{document}